# Location-routing Optimisation for Urban Logistics Using Mobile Parcel Locker Based on Hybrid Q-Learning Algorithm

**Yubin Liu**
Department of Civil and Environmental Engineering
Imperial College London, London, UK, SW7 2AZ
Email: y.liu20@imperial.ac.uk

**Qiming Ye**
Department of Civil and Environmental Engineering
Imperial College London, London, UK, SW7 2AZ
Email: qiming.ye18@imperial.ac.uk

**Yuxiang Feng**
Department of Civil and Environmental Engineering
Imperial College London, London, UK, SW7 2AZ
Email: y.feng19@imperial.ac.uk

**Jose Escribano-Macias**
Department of Civil and Environmental Engineering
Imperial College London, London, UK, SW7 2AZ
Email: jose.escribano-macias11@imperial.ac.uk

**Panagiotis Angeloudis**
Department of Civil and Environmental Engineering
Imperial College London, London, UK, SW7 2AZ
Email: p.angeloudis@imperial.ac.uk

Word Count: 7,229 words + 250 (1 tables) = 7,479 words

*Submitted 1 August 2021*

**ABSTRACT**

Mobile parcel lockers (MPLs) have been recently introduced by urban logistics operators as a means to reduce traffic congestion and operational cost. Their capability to relocate their position during the day has the potential to improve customer accessibility and convenience (if deployed and planned accordingly), allowing customers to collect parcels at their preferred time among one of the multiple locations. This paper proposes an integer programming model to solve the Location Routing Problem for MPLs to determine the optimal configuration and locker routes. In solving this model, a Hybrid Q-Learning algorithm-based Method (HQM) integrated with global and local search mechanisms is developed, the performance of which is examined for different problem sizes and benchmarked with genetic algorithms. Furthermore, we introduced two route adjustment strategies to resolve stochastic events that may cause delays. The results show that HQM achieves 443.41% improvement on average in solution improvement, compared with the 94.91% improvement of heuristic counterparts, suggesting HQM enables a more efficient search for better solutions. Finally, we identify critical factors that contribute to service delays and investigate their effects.

## INTRODUCTION

The growth of e-commerce stimulates the demand for timely parcel delivery and collection, contributing to intensive logistics operations in urban areas. However, the increasing logistics activity had led to adverse effects on urban mobility, such as limited parking space and traffic congestion [1]. To address these concerns, novel last-mile concepts have been proposed to improve convenience and efficiency, such as deploying mobile parcel lockers (MPLs). These can facilitate flexible delivery according to customers' location and increase accessibility for customers while fewer lockers are required [2].

The range of algorithms that have been previously used to solve the mobile parcel locker problem (MPLP) includes exact algorithms [3], classic heuristics [2], and meta-heuristics [4]. However, a key disadvantage of implementing heuristic algorithms is that they are vulnerable to the presence of local optimums when large instances are deployed—an issue that this study attempts to improve.

The motivation for this study is twofold. First, different studies on MPLP have identified the lockers' location as the most critical factor that influences customers' acceptance using locker services [5, 6]. As a result, the effectiveness of stationary parcel lockers is limited by the spatial variability in customer demand. Hence, deploying mobile lockers that consider dynamic customer locations has the potential of enabling couriers to respond to demand changes in a more agile manner. The second motivation is the potential of our adopted solution approach (Q-Learning) to improve the quality of solutions given its ability to consider a wide spectrum of individual behaviour [7]. Compared with meta-heuristic algorithms, Q-Learning can better utilise the information feedback obtained from previous actions, and is capable of improving previously determined solutions while retaining the elements that contribute to its determined effectiveness.

Given these considerations, this study pursues the following objectives: 1) To develop task and route generation algorithms to define the task sequence considering customers' locations and time windows constraints, 2) To develop a Hybrid Q-Learning algorithm-based Method (HQM) resolving the disadvantage of falling into local optimum for large-sized networks faced by most evolutionary algorithms, and 3) To evaluate the performance of HQM comparing to a meta-heuristic algorithm (Genetic Algorithm) using simulated data and define the effect of critical factors against service delay.

This paper contributes to the existing literature in the following respects. First, the proposed HQM enables deriving better solutions than meta-heuristic algorithms for solving large-scale MPLP problems. Second, two route adjustment strategies are developed to dynamically resolve the time windows conflicts caused by service delay and traffic conditions. Third, our analysis reveals that HQM is an effective framework to solve the MPLP regardless of the network type. This is beneficial to customers and express companies, allowing them to handle parcel deliveries' timeliness and accessibility problems in a coordinated and flexible fashion. Finally, this research enriches the application of the Q-Learning method in the field of urban logistics optimisation.

The rest of the paper is structured as follows. It begins with a brief review of parcel locker-related studies. The problem statement introduces the MPLP and formulates an integer programming model. The methodology section presents our proposed solution approach for the MPLP instances, consisting of three parts: i) task generation procedure, ii) route generation and adjustment procedure, and iii) HQM method integrated with global and local search. The results and discussion section discusses the results of our experiments and identifies the effects of critical factors against service delay. Finally, the conclusion provides a summary of findings and highlights the potential application of our MPLP solution approaches.

## LITERATURE REVIEW

The research of optimising last-mile deliveries has expanded substantially in recent years. For the purposes of this review, we focus on more recent papers on novel trends in delivery with autonomous vehicles [8], pickup stations [9], and e-fulfilment with shared reception boxes [10].

To the best of the authors' knowledge, literature on location-routing planning for MPLs is limited: only four pieces of literature focus on the last-mile deliveries using MPLs [2, 3, 4, 11]. Since the MPLP involves aspects of facility location selection and vehicle route planning, we analyse the following fields related to our research: (i) MPLP, (ii) location-routing problems (LRPs), and (iii) vehicle routing problem (VRPs) with reinforcement learning (RL).

*(i) MPLP:* Schwerdfeger and Boysen [3] introduced the first MPLP, the formulation which bears a significant degree of similarity with our proposed model. A mixed-integer programming model was formulated to minimise the fleet size of MPLs while covering all customers through MPLs relocation. While their exact solution approach shows substantial improvements compared to stationary locker systems, several key limitations arise. Firstly, there is a lack of consideration of stochastic factors (e.g., service time) in the study, which affect the dynamic adaption of the scheduling strategy. Moreover, the operational details of a single mobile locker cannot be tracked, such as the service starting/ending time and the duration at parking spaces. Those operational details have a significant impact on improving scheduling policies and customer satisfaction. Orenstein et al. [11] introduced a flexible parcel delivery problem that enables customers to provide couriers with multiple delivery locations; each parcel will only be delivered to a subset of the locker network. The solution approach is based on a savings heuristic, the petal method, combined with tabu search with a joint objective of large neighbourhoods to optimise the delivery task with lower cost and shorter delivery times. The results show that the proposed method could be adapted to dynamic and stochastic environments. However, the model was not adapted to the occasions where customers may change their prefered destinations if parcels fail to be delivered. This limitation is worth discussing since customers may request delivery again, which requires the model to respond dynamically to requirements.

Wang et al. [4] integrated the LRPs with MPLs, and established a non-linear integer programming model to minimise operating costs. An embedded GA was developed to determine the locations of depots, the number of mobile parcel lockers deployed, and routes of lockers simultaneously. The results show that a policy that considers demand aggregation can significantly reduce delivery time while the scheme without demand aggregation saves the number of vehicles deployed. Li et al. [2] introduced a Two-Echelon MPLP, in which locker travel to couriers in the field and transfer parcels between the couriers and the depot, while couriers perform the final delivery without making repeated returns to the depot. A hybrid Clark and Wright heuristic was developed to solve the mixed-integer programming model. However, the proposed method becomes computationally infeasible for larger problem instances. Both these studies lack the consideration of dynamic cases, where locker routes are adjusted to respond to stochastic events (e.g., demand changes and traffic congestions).

This study will resolve the limitation mentioned above. Concretely, the dynamic route adjustment for stochastic events and the computing complexity for large network sizes. The HQM embedded with route adjustment strategies is developed to solve these limitations.

*(ii) LRPs* combine two decision tasks: (1) the location of varying types of facilities, and (2) the allocation scheme in which customer fulfilled by which facility and the corresponding delivery route. The common objectives of LRPs include minimising the total travel cost, the acquisition of vehicles, and the cost of locating and operating depots [12, 13].

Most literature studying multi-echelon LRPs focuses on two-echelon cases and ignores temporal aspects such as time windows or the synchronisation of transshipments [14, 15]. The heuristics proposed by Perboli et al. [14] and Gonzalez Feliu et al. [15] indicate the fact that once a customer is assigned to a facility at the previous echelon, the problem can be decomposed into a single VRP for a facility at a higher echelon and another VRP for each facility at a sub-echelon.

Our proposed model is a single-echelon LRPs model. Although the echelon of the customer-to-locker network can be regarded as the second echelon of the model, we do not plan the route between customers and MPLs. To the best of the authors' knowledge, this aspect has not been treated within the area of two-echelon LRPs. Instead, we determine the locker routes from the central depot towards parking spaces and the routes between parking spaces.

*(iii) VRPs with RL:* VRPs are combinatorial optimisation problems that are computationally expensive. Reinforcement learning provides a compelling choice to resolve the computational complexities faced by classic heuristic algorithms in the pursuit of an optimal solution. The advantage of RL lies in its learning ability as it interacts with the environment, thus improving the solution more efficiently. The optimal solution can be regarded as a sequence of decisions in RL following the Markov Decision Process (MDP) of the problem. Hence, decision-makers can apply RL to obtain near-optimal solutions by increasing the probability of decoding expected sequences [16]. Bello et al. [17] were the first to apply policy gradient algorithms to Traveling Salesman Problem (TSP). The MDP can be represented as follows: a state is denoted as a $p$-dimensional graph embedding vector, representing the current tour of the nodes at time step $t$. The action is whether to select another unselected node or not for the current state. The transition function $T(s, a, s')$ returns the next node from a set of the constructed tour until all nodes have been visited. The reward function is the negative tour length.

Nazari et al. [18] proposed a VRP model, in which state $s$ is represented as a vector of tuples, including customer's location and demands. The action represents whether to add a node to the current route or not, such that vehicles will visit. The reward is denoted as the negative value of the total travel distance. Since then, hybrid RL were gradually used to solve VRP.

Cappart et al. [19] combined RL and constraint programming (CP) to solve VRP with time windows. Deep Q-Networks and Proximal Policy Optimization algorithms were trained to formulate MDP to determine efficient branching policies for different CP search strategies. Yu et al. [20] developed a deep RL-based neural combinatorial optimisation strategy to transform online VRP into a vehicle route generation problem. They proposed a structural graph embedded pointer network to develop vehicle routes. A deep RL mechanism based on an unsupervised auxiliary network was used to train model parameters. The result shows that the proposed strategy achieves a fast online route generation speed due to the offline parameter training process. Zhao et al. [21] proposed a novel deep reinforcement learning (DRL) model, which composes an actor, an adaptive critic, and a routing simulator. The actor is designed to generate routing strategies based on attention mechanism. An adaptive critic is developed to adjust the network structure accordingly such that improving solution. The output of the DRL model is used as the initial solution of the following local search method from which the final solution can be obtained.

Since our model does not rely on historical training samples and prior knowledge, states are also generated randomly according to probability distributions, we propose a novel hybrid RL method that integrates Q-Learning with global and local search mechanisms to solve MPLP.

## PROBLEM STATEMENT

A set of customer nodes $N$ serve as the recipients of parcels from a group of MPLs. For each customer $n \in N$, customer $n$ creates a set of location $L_n$ during the planning horizon. Each location that customer visited within the location set ($k \in \{1, 2, \cdots, \|L_n\|\}$) is defined by a time interval $[E_{nk}, L_{nk}]$ and a corresponding position $[X_{nk}, Y_{nk}]$.

Customer $n$ will pick up a parcel from a specific parcel locker if the locker's current stopover is within the customer's maximum walking range $\rho_n$. We assume that the maximum accepted walking ranges $\rho_n$ can vary from customer to customer and can be changed at different locations (e.g., a customer tends to have a higher possibility to decline large $\rho_n$ after work).

A set of mobile lockers $M$ change their locations to facilitate parcel pickups by the customer set $N$. Since MPLs cannot park randomly in urban areas, we predefine a given set of parking spaces $I$ where mobile locker $m$ can potentially be parked. For each parking space $i \in I$, we predefine a position $[X_i, Y_i]$, and the available parking time windows $[E_i, L_i]$. We define $S_i$ to denote the service time of parking space $i$. Therefore, for each parking space $i$, the parking time windows $[E_i, L_i]$ can be divided into $u_i$ ($u_i = |L_i - E_i|/S_i$) sub-intervals. Each sub-interval has an equal length of time span. We introduce $[e_{ia}, l_{ia}]$ to represent the $a$th ($a = 1, \dots, u_i$) sub-interval within $[E_i, L_i]$ in parking space $i$. The demand within each sub-interval $[e_{ia}, l_{ia}]$ to be fulfilled is define as a single task. To avoid missed deliveries (e.g., due to traffic congestion) and guarantee feasible parcel handovers at selected stopovers, we define a buffer time $\delta$ to denote the minimum overlap between $[e_{ia}, l_{ia}]$ and $[E_{nk}, L_{nk}]$.

Figure 1 illustrates an example of MPLP with four customers and one parcel locker. The four customers are denoted by the solid circle with green, purple, orange, and yellow, and their corresponding pickup-interval time are represented as a blue rectangle. Customers may change location throughout the day as denoted with the dotted coloured arrows. Note that the customer time windows at different locations shall never overlap. The proposed MPLP aims to allocate the least number of MPLs departed from depots, fulfil all customers at a specific time interval within a day, and plan locker routes. The optimal solution here is deploying one MPL to satisfy all customers, as shown by the solid black arrow (since the available parking time intervals satisfy both customer time windows and service buffer time).

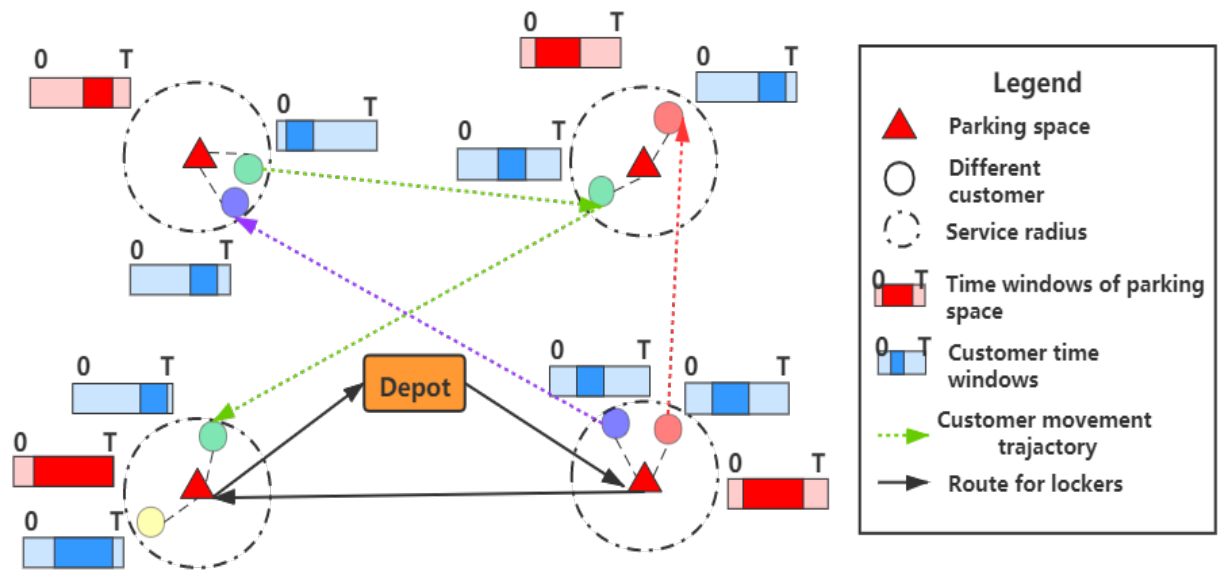

**FIGURE 1** Instance of MPLP

We aim to minimise cost, which consists of the number of lockers deployed and the total distance travelled while satisfying all customers and time windows constraints from both customer and parking space. The notation and the mathematical model are as follows:

Indices:

| | |
|---|---|
| $m$ | mobile parcel locker $m$ ($m \in M$) |
| $n$ | customer $n$ ($n \in N$) |
| $i, j$ | two adjacent parking spaces ($i, j \in I$) |

Sets:

| | |
|---|---|
| $N$ | set of customer nodes |
| $M$ | set of MPLs |
| $I \backslash \{0\}$ | set of parking space, where 0 represent depot |
| $V$ | set of nodes within the network ($V = \{0, 1, \dots, \|I\|\}$ ), where 0 represent depot |

Parameters:

| | |
|---|---|
| $Q$ | the capacity of MPLs |
| $q_{ia}$ | the total demand to be fulfilled within the $a$th sub-interval of parking space $i$ |
| $[e_{ia}, l_{ia}]$ | the $a$th sub-interval within $[E_i, L_i]$ at parking space $i$ ($a = 1, \dots \|u_i\|$) |
| $t_{ij}$ | time consumption travelling from $i$ to $j$ |
| $S_i$ | service time of parking space $i$ |
| $A_{ima}$ | the time when locker $m$ arrives at the $a$th sub-interval of parking space $i$ |
| $F_m$ | fixed cost of locker $m$ |
| $C_{ij}$ | unit cost travelling from parking space $i$ to $j$ |
| $\mu$ | a large indefinite number |

Decision Variables:

| | |
|---|---|
| $x_{ijm}$ | Boolean: 1, if locker $m$ visits parking space $j$ after fulfill the demands at $i$; 0, otherwise |
| $y_{im}$ | Boolean: 1, if locker $m$ serves the customer covered by parking space $i$; 0, otherwise |
| $z_{ima}$ | Boolean: 1, if locker $m$ starts service within the $a$th sub-interval of parking space $i$; 0, otherwise |

**Mathematical Model**

$$Minimise \; f(x_{ijm}, y_{im}, z_{ima}) = W_1 \times F_m \times \sum_{m=1}^{M} \sum_{i=1}^{I} x_{i0m} + W_2 \times \sum_{m=1}^{M} \sum_{i=0}^{I} \sum_{j=0}^{I} x_{ijm} \times C_{ij} \quad (1)$$

*Subject to:*

$$\sum_{m=1}^{M} \sum_{i=0}^{I} x_{ijm} = 1 \quad \forall i \in I \quad (2)$$

$$\sum_{m=1}^{M} \sum_{j=0}^{I} x_{ijm} = 1 \quad \forall j \in I \quad (3)$$

$$\sum_{i=1}^{I} x_{i0m} = 1 \quad \forall m \in M \tag{4}$$

$$\sum_{j=1}^{I} x_{0jm} = 1 \quad \forall m \in M \tag{5}$$

$$\sum_{i=1}^{I} \sum_{a}^{u_i} q_{ia} \, y_{im} \le Q \quad \forall m \in M \tag{6}$$

$$e_{ia} - \mu(1 - z_{ima}) \le A_{ima} \quad \forall m \in M; i \in I; a \in \{1, \dots \|u_i\|\} \tag{7}$$

$$l_{ia} - \mu(1 - z_{ima}) \ge A_{ima} \quad \forall m \in M; i \in I; a \in \{1, \dots \|u_i\|\} \tag{8}$$

$$A_{ima} + S_i x_{ijm} + t_{ij} x_{ijm} - \mu(1 - x_{ijm}) \le A_{jma} \quad \forall m \in M; \ i,j \in I \tag{9}$$

$$e_{ja} - \mu(1 - z_{jma}) \le A_{ima} + S_i x_{ijm} + t_{ij} x_{ijm} \quad \forall m \in M, i \in I \tag{10}$$

$$l_{ja} - \mu(1 - z_{jma}) \ge A_{ima} + S_i x_{ijm} + t_{ij} x_{ijm} \quad \forall m \in M, i \in I \tag{11}$$

$$x_{ijm}, y_{im}, z_{ima} \in \{0,1\} \quad \forall m \in M; i,j \in I \tag{12}$$

The objective function (1) minimises the fleet size and the total travel distance, where $W_1$ and $W_2$ represent the weights of two cost units. Constraints (2) and (3) ensure that parking space $i$ and $j$ can only be visited once within the same sub-interval. Constraints (4) and (5) guarantee MPLs depart from a central depot and return when the task is completed. Constraint (6) represents that the customer fulfilled by locker $m$ will not exceed the locker's capacity. Constraints (7) and (8) ensure that the locker $m$ arrives at parking space $i$ within the $a$th sub-interval if locker serves the corresponding customers. Constraints (9) to (11) represents that if locker $m$ visits parking space $j$ after fulfilling the customer at $i$, the time arriving at $j$ needs to meet the time window of $j$. Constraint (12) defines the domain of the binary variables.

## METHODOLOGY

The structure of our proposed solution approach is illustrated in Figure 2. First, we assign different customer locations to the nearest parking space using K-means clustering, such that MPLs can fulfil customers at one of his/her potential locations within a day. Next, we derive a finite set of tasks for each parking space to compose the original problem into defining task allocation and the sequence of executing tasks by different lockers. The task set of parking space $i$ is constituted by multiple sub-tasks within $[E_i, L_i]$. Notably, the sub-task is defined as the demand to be fulfilled within each sub-interval $[e_{ia}, l_{ia}]$ (see Problem statement). MPLs are then allocated to perform the task set of all parking spaces.

By obtaining a finite set of tasks, the HQM algorithm is designed to determine the following set of decisions: i) task allocation schemes (denoted as $x_1$) that determine sub-task and locker pairings, and ii) the locker task execution sequence (referred as $x_2$). A state $s = (x_1, x_2)$ can be regarded as a feasible solution of the proposed MPLP. An agent is regarded as a set of state $s$, and is used to accelerate the search for an optimal solution.

Randomly generated agents serve as inputs to the HQM, which generates an optimal state of each agent based on its current knowledge of the problem. Thus, the task allocation scheme and the task execution sequence are determined. The sequence of tasks performed by different MPLs determines the locker routes. The essential elements and structure of the HQM approach will be demonstrated in the following sections.

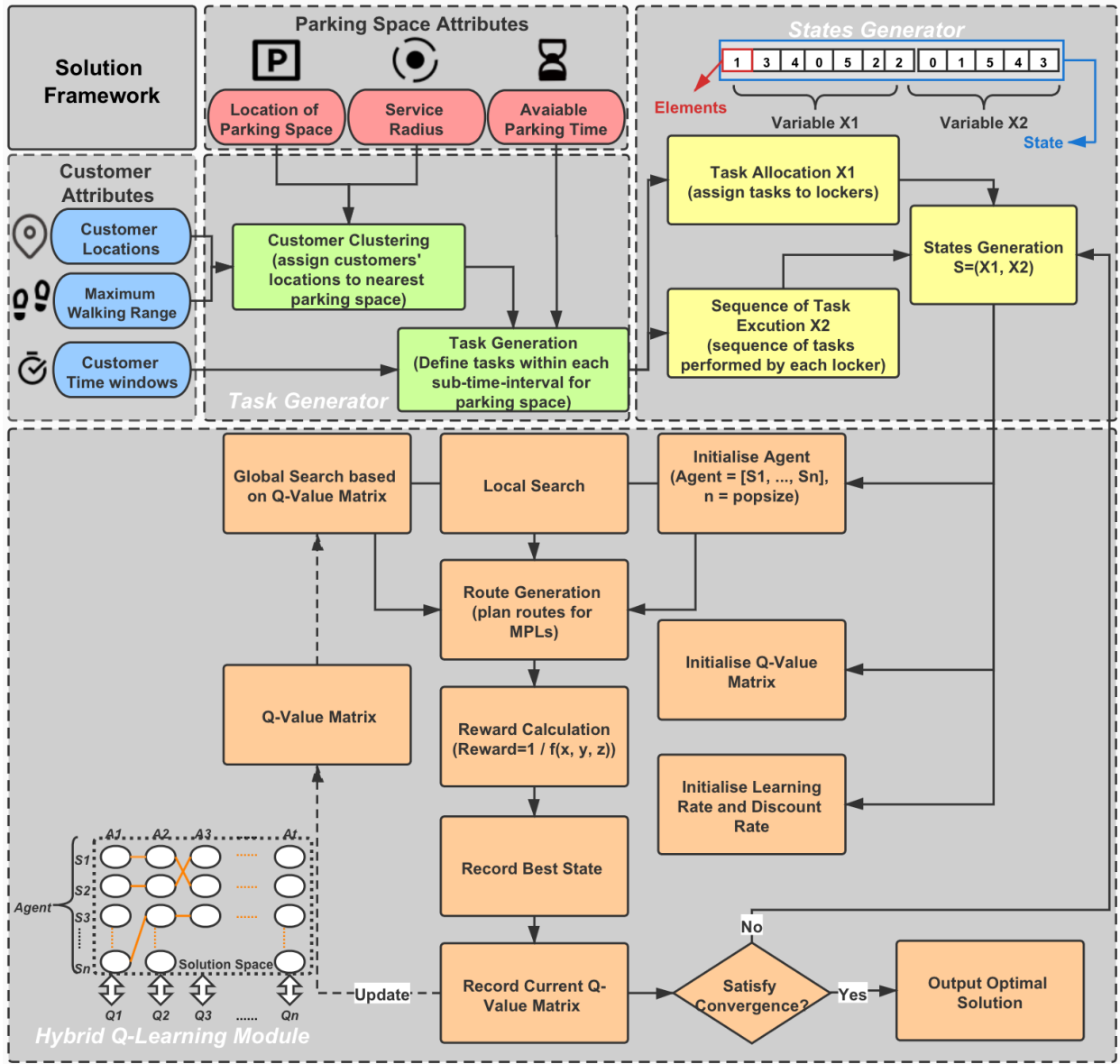

**FIGURE 2** Solution Framework of Model

## Task Generation

Any planning horizon and availability interval of each parking space can be divided into a finite set of sub-intervals once a service time $S_i$ of parking space has been defined (see Problem statement). We first reduce the redundant length of parking spaces' availability intervals. For instance, if a parking space $i$ is available for a mobile locker from 10:00 to 16:00, but no customer time windows exist between 14:00 and 16:00, we can reduce the available time intervals to only last from 10:00 to 14:00 without excluding optimal solutions. The available time interval of parking space $i$ can be reduced to $U_i$:

$$U_i = \{[E_i, L_i] \cap [E_{nk}, L_{nk}] \ \forall n \in N; \ k \in \{1, 2, \cdots, \|L_n\|\}; \ i \in I\} \tag{13}$$

To further reduce the range of available time intervals, we determine the earliest time window and the latest time window among all customers within each parking space. The optimal range of available time intervals will be updated as $U_i^{opt}$. By defining $U_i^{opt}$, we obtain a set of equal-length sub-intervals (Equation 15) of parking space $i$ when executing $U_i^{opt}$ divided by service time $S_i$

$$U_i^{opt} = \left\{ [min\{E_{1k}, \cdots, E_{nk}\}, max\{L_{1k}, \cdots, L_{nk}\}] \cap U_i \ \forall n \in N; k \in \{1, \cdots, \|L_n\|\} \right\} \tag{14}$$

$$[e_{ia}, l_{ia}] = U_i^{opt} / S_i \ \ \forall i \in I; a \in \{1, 2, \cdots, u_i\}; \ [e_{ia}, l_{ia}] \in U_i^{opt} \tag{15}$$

Thus, the set of demands to be fulfilled by mobile locker $m$ at parking space $i$ within the $a$th sub-interval can be denoted by $Q_{ima}$, where $q_{ia}$ denotes the total demand at the $a$th sub-interval at $i$. Each $Q_{ima}$ is regarded as a single task. The task list of locker $m$ is denoted as $TK_m = \{Q_{1ma}, \cdots, Q_{ima}\}$.

$$Q_{ima} = \{(q_{ia}, m, i, [e_{ia}, l_{ia}]) \ \ \forall i \in I; \ m \in M; \ a \in \{1, 2, \cdots, u_i\} \} \tag{16}$$

**Route Generation**

*Mechanism of Route Generation*

We propose a route generation algorithm to define the sequence of mobile lockers executing the tasks generated in *Task Generation*. The generated routes will be applied in the hybrid Q-Learning model in the later section; The evaluation of the generated route is measured by Equation (1). For each parking space, it consists of multiple tasks within the available parking time intervals. The locker will perform the corresponding task within the sub-intervals. Considering the optimal scheduling strategy, a locker may also visit the same parking space multiple times within different sub-intervals to perform corresponding tasks. To capture the route generation process of mobile lockers, we modified a lemma developed by Schwerdfeger and Boysen [3] and prove that for any given task $Q_{ima}$, the task starting/ending time can be adjusted to satisfy constraints (7) to (11).

**Lemma 1.** *For any $Q_{ima}$ ($i \in I, a \in \{1, 2, \cdots, u_i\}$), there is an optimal task starting and ending time to fulfil the corresponding customer/demand, while constraints (7) to (11) are satisfied.*

We assume that a locker arrives at its first parking space at the earliest time within $[e_{1a}, l_{1a}]$ of $Q_{1ma}$ and leave it once the last task $Q_{ima}$ is fulfilled by the earliest time. Then, the locker moves to the next parking space and starts executing the task when it arrives (if the arriving time $A_{ima}$ is within the corresponding sub-interval of the next task). As such, the locker leaves the current parking space and moves towards the next one earlier. Following this logic, Lemma 1 can be proven.

**Proof** To simplify the process, we assume that each customer only has a single location ($L_n = 1$), but our proof is still valid for the multiple-locations situation. Following the symbols defined in problem statement section, we define $A_{ima}$ as the time when locker $m$ arrives at the $a$th sub-interval of parking space $i$. $\overrightarrow{A_{1ma}}$ is the leaving time, $S_i$ is the service time of parking space $i$, and $t_{ij}$ is the travelling time from $i$ to $j$. At the first parking space $i$ of locker $m$, we can set:

$$A_{1ma} = \max\{e_{1a}, t_{01}\} \ \forall m \in M; a \in \{1, \cdots, u_i\}; \ e_{1a} \in [e_{1a}, l_{1a}] \tag{17}$$

We assume that the locker serves the customers as early as possible, because a task can start immediately at the beginning of the sub-interval. Thus, the locker can leave at the earliest at:

$$\overrightarrow{A_{1ma}} = \{\max(A_{1ma}, E_{nk}) + S_1\} \ \forall m \in M; n \in N; a \in \{1, \cdots, u_i\}; \ E_{nk} \in [E_{nk}, L_{nk}] \tag{18}$$

because the last customer is served as soon as possible. $E_{nk}$ denotes the earliest time among all customers within the current sub-interval. The earliest task starting time of the second parking space is:

$$A_{2ma} = \max\{\overrightarrow{A_{1ma}} + t_{12}, e_{2a}\} \ \forall m \in M; a \in \{1, \cdots, u_i\}, e_{2a} \in [e_{2a}, l_{2a}] \tag{19}$$

Therefore, the earliest task starting and ending time for each $Q_{ima}$ can be represented as:

$$A_{ima} = \max\{\overrightarrow{A_{(i-1)ma}} + t_{(i-1)i}, e_{ia}\} \ \forall m \in M; a \in \{1, \cdots, u_i\}, e_{ia} \in [e_{ia}, l_{ia}] \tag{20}$$

$$\overrightarrow{A_{ima}} = \max\{\max\{A_{ima}, E_{nk}\} + S_i \ \forall m \in M; n \in N; a \in \{1, \cdots, u_i\}; \ E_{nk} \in [E_{nk}, L_{nk}] \tag{21}$$

*Route adjustment strategy*
The current load of the locker may reach its maximum capacity before performing the following task and violating constraint (6), meaning that the locker must return to the depot to unload and then execute the next task. Additionally, the task ending time ($A_{ima} + S_i$) within the $a$th sub-interval of parking space $i$ plus the travelling time $t_{ij}$ towards parking space $j$ may less than the earliest time $e_{ja}$ of $j$ due to the stochastic driving time (violate constraints (7) to (11)). We propose two strategies to postpone the arrival time to satisfy constraints: 1) the locker moves to the depot and then visits the next parking space (refer as BTD); 2) the locker remains at the current parking space until the latest ending time $l_{ia}$ of the $a$th sub-interval (refer as HCPS). The following relationships capture the situation where a locker travels from parking space $i$ to $j$ using the above adaption strategies:

$$A_{jma} = \begin{cases} A_{ima} + S_i + t_{i0} + t_{0j} & \forall i, j \in I; m \in M; a \in \{1, 2, \cdots, u_i\} \tag{22} \\ l_{ia} + t_{0j} & \forall i, j \in I; m \in M; a \in \{1, 2, \cdots, u_i\} \tag{23} \end{cases}$$

Equation (22) represents the task starting time adapted by implementing BTD, which equals the previous task ending time plus the time consumption travelling between the depot and two parking spaces. Equation (23) represents the situation implementing HCPS, which equals the latest ending time of the previous task plus the time consumption travelling to the next parking space. By using different strategies, the starting time of the next task satisfies $e_{ja} < A_{jma} < l_{ja}$. Code Listing 1 outlines the task and route generation process and route adjustment procedure.

---

**Algorithm 1:** Route generation and adjustment algorithm

---

Input: Number of lockers $M$; Task list $TK_m$ for $m \in M$; Locker speed $v$
output: The reward of the generated routes

1 *Initialise corresponding parameters and matrix*
2 Calculate travel time $t_{ij}$ between parking space $i$ and $j$
3 **for** each locker $m$ *in* $M$ **do**
4      index = argsort($TK_m$)
5      $TK_m = TK_m$[index]
6      total_dispatch += 1 // Record the dispatched locker
7 **end**
8 **for** each sub-task $tk$ *in* $TK_m$ **do**
9      Add depot to current task list $TK_m$ // Depart from depot
10      Create blank list *route_list* // Record visited parking spaces
11      Initialise $A_{ima}, q_{ia}, e_{ia}, l_{ia}, E_i, L_i, S_i$
12      **for** *next* **in** $tk$ **do**
13          current_parking_space = $TK_m[tk]$
14          next_parking_space = $K_m[next]$
15          Add current_parking_space to *route_list*
16          Update $A_{ima}, q_{ia}, e_{ia}, l_{ia}$
17          travel_time = lialg.norm(current_parking_space − next_parking_space) / $v$
18          **if** $A_{ima} + S_i + t_{ij} < e_{ja}$ **and** $l_{ja} + t_{ij} \geq e_{ja}$ **then** // Violate constraints (7) to (11)
19              Adjust task starting and ending time based on Equation (23) // Apply HCPS
20          **else if** $A_{jma} < e_{ja}$ **or** $q_{ia} > Q$ **then** // Violate constraints (7) to (11) or constraint (6)
21              Adjust task starting and ending time based on Equation (22) // Apply BTD
22              Update $A_{jma}, q_{ja}, e_{ja}, l_{ja}, E_j, L_j, S_j$
23              Add next_parking_space to *route_list*
24          **else if** *satisfy constraints* **then**
25              Update $A_{jma}, q_{ja}, e_{ja}, l_{ja}, E_j, L_j, S_j$
26              Add next_parking_space to *route_list*
27          **end**
28      **end**
29 **end**
30 Calculate the total route length from *route_list*
31 Calculate the value of objective function $f(x_{ijm}, y_{im}, z_{ima})$
32 Reward $\leftarrow 1/f(x_{ijm}, y_{im}, z_{ima})$
33 **return** *Reward*

---

**CODE LISTING 1** Route Generation and Adjustment Algorithm

**Hybrid Q-Learning Algorithm-based Model (HQM)**

Code Listing 1 obtains locker routes that can be used to execute delivery tasks, and calculates the corresponding reward. We further introduce a hybrid Q-Learning model that combines global and local search to generate a set of feasible solutions and obtains the optimal result by updating Q-Value iteratively.

Given $\pi$ as the integration of task allocation scheme and the tasks execution sequence, our goal is to maximise the reward $R$ obtained from a set of generated locker route $G$:

$$R(\pi|G) = \max \left\{ \frac{1}{f_{\pi_1}(x_{ijm}, y_{im}, z_{ima})}, \cdots, \frac{1}{f_{\pi_n}(x_{ijm}, y_{im}, z_{ima})} \right\} \quad \pi_n \in \|Agent\| \quad (24)$$

The proposed HQM, parameterised by $\theta_1$ (correspond to task allocation scheme $x_1$) and $\theta_2$ (correspond to task executing sequence $x_2$), needs to define a strategy $p(\pi|G)$ that improve the solution $\pi$. It is a stochastic policy and can be factorised as:

$$p_{\theta_1, \theta_2}(\pi|G) = \prod_{i=1}^{t} p_{\theta_1, \theta_2}(\pi(t)|\pi(<t), G) \qquad (25)$$

To maximise the objective in Equation (24), we use HQM to update every component on the right-hand side of Equation (25) in every timestep $t$. The essential elements of HQM are defined as follows:

*A. State*

State $s = (x_1, x_2)$ is determined as a vector, consisting of two variables: 1) the task allocation scheme ($x_1$) that defines which task is assigned to which locker, and 2) the task execution sequence of MPLs ($x_2$). Each state is regarded as a feasible solution, and different states form an agent. An illustration is shown in the upper right of Figure 2.

*B. Agent*

We consider a set of states $s_n = (x_1, x_2)$ as an agent ($Agent = \{s_1, \cdots, s_n\}$). At every timestep $t$, the agent interacts with the environment and selects an action $\pi(t)$ based on its policy. By selecting an action $\pi(t)$, the state within the agent will be updated accordingly based on the Q-Value at every timestep $t$. The larger the reward a state obtains, the larger the Q-Value it has. The new state corresponding to the agent generated in the next timestep/action will be similar to the current state with the maximum Q-value in the current timestep/action, meaning that the new state retains the elements that make a significant contribution to the optimal solution in the current timestep. We generate multiple agents for the later global search to speed up the search for the optimal solution. Receiving a new reward $R$ from the environment, the agents are iteratively updated with improved states.

*C. Action*

The action $\pi(t)$ can be defined as a decisional behaviour that agents select a policy searching for a better solution in the next timestep. $\pi(t)$ is sampled from the right-hand side of Equation (25) during training time, and obtained by $\varepsilon - greedy$ search of the Q-Learning and global-local search strategies.

*D. Reward*

The reciprocal of Equation (1) is calculated as a reward when all tasks are executed, since the HQM algorithm is set to maximise the reward (minimise Equation (1)).

**Decomposition of State-Action Space**

A lookup table represents the state-action space of HQM. The size of the lookup table equals the Cartesian product of $S \times A$, where $S$ equals the length of elements (feasible solution) within the agent, while $A$ equals the number of iteration (step size). Hence, the size of the lookup table will expand exponentially when the problem size becomes larger, causing computing complexity.

To tackle such a curse of dimensionality, we decomposed the state-action space into multiple low-dimensional state-action combination chains linked by Q-value matrix; we define two Q-Value matrices, which correspond to the variable $x_1$ and $x_2$ within a state respectively. The two Q-Value matrices are captured by Equations (26) and (27), where $m$ equals the number of lockers deployed and $n$ equals the number of tasks. Each matrix value reflects the evaluation of the variable, such that the current task allocation scheme and the task executing sequence can be evaluated. Additionally, each value within the matrix reflects the interdependence between two adjacent actions. The larger the value is, the more closely the two adjacent actions are connected. The new state corresponding to the agent updated in the next action will inherit the elite fragment of the state with the larger Q-Value in the current action. Therefore, it is possible to improve the solution while inheriting an elite fragment of the current state.

$$p(\theta_1) = \begin{bmatrix} Q_{11}(\theta_1) & \cdots & Q_{1m}(\theta_1) \\ \vdots & \ddots & \vdots \\ Q_{m1}(\theta_1) & \cdots & Q_{mm}(\theta_1) \end{bmatrix} \tag{26}$$

$$p(\theta_2) = \begin{bmatrix} Q_{11}(\theta_2) & \cdots & Q_{1n}(\theta_2) \\ \vdots & \ddots & \vdots \\ Q_{n1}(\theta_2) & \cdots & Q_{nn}(\theta_2) \end{bmatrix} \tag{27}$$

The iteration procedure can be regarded as an MDP that derives satisfactory results through updating the Q-value between two adjacent actions. Figure 3 shows the structure of the $S \times A$ dimension lookup table; For each action $A_i$ ($i \in \{1, \cdots, t\}$), there is a corresponding Q-Value ($p(\theta_1)$ and $p(\theta_2)$) that interacts with it. Each action selects the direction of migration based on the Q-Value: the action $A_i$ is taken as the previous iteration of the next action $A_{i+1}$ once $A_i$ is determined, and the next action will be selected according to $Q_{i+1}$.

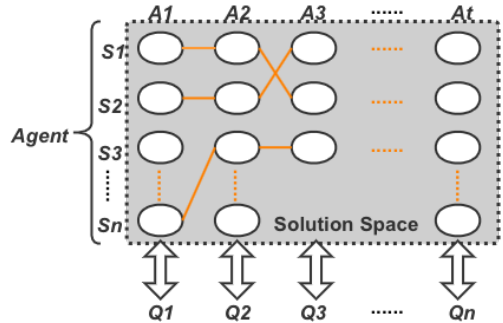

**FIGURE 3** State-Action Space

The update of the two Q-Value matrices follows the following equation:

$$Q_{t+1}^{x_i}\left(s_t^{x_{ij}}, a_t^{x_{ij}}\right) = Q_t^{x_i}\left(s_t^{x_{ij}}, a_t^{x_{ij}}\right) + \alpha \left[ R^{x_{ij}}\left(s_t^{x_{ij}}, s_{t+1}^{x_{ij}}, a_t^{x_{ij}}\right) + \gamma \max_{a^{x_{ij}} \in A_t} Q_t^{x_i}\left(s_{t+1}^{x_{ij}}, a^{x_{ij}}\right) - Q_t^{x_i}\left(s_t^{x_{ij}}, a_t^{x_{ij}}\right) \right] \tag{28}$$

where $x_i$ represents the type of variable within the state ($x_i \in \{x_1, x_2\}$), $j$ represent the $j$th agent, $R^{x_ij}\left(s_t^{x_ij}, s_{t+1}^{x_ij}, a_t^{x_ij}\right)$ represents the reward obtained from state $s_t^{x_ij}$ executing action $a^{x_ij}$ towards $s_{t+1}^{x_ij}$ in timestep $t$, $\alpha$ represents the learning rate, and $\gamma$ represents the discount factor.

**Policy for Action Selection**

The trade-off between exploitation and exploration of the information feedback obtained from previous actions is challenging for traditional Q-Learning. Emphasising information exploration may obtain the optimal solution but sacrifice convergence speed. While emphasising information exploitation improves convergence speed at the cost of optimality. We propose to integrate the Q-Value-based global search with a local search approach to define the policy for action selection, such that obtains a near-optimal solution while improves convergence speed.

*Global Search Based on Q-Value Matrix*

The transfer between two adjacent actions updates agents according to the Q-Value obtained from the last action. Since the new agents consist of multiple new states, each composed of variables $x_1$ and $x_2$, we first apply a global search to produce elements for each state variable parameter. The policy of action selection is based on the $\varepsilon - greedy$ search which can be defined as follows:

$$a_{t+1}^{x_ij} = \begin{cases} \underset{a^{x_ij} \in A_t}{\operatorname{argmax}} Q^{x_i}\left(s_{t+1}^{x_ij}, a^{x_ij}\right), & \sigma < \varepsilon \\ a^{x_ij\prime}, & \sigma \geq \varepsilon \end{cases} \tag{29}$$

Where $\varepsilon$ denotes the greedy factor, $\sigma$ denotes the random value between 0 and 1, and $a^{x_ij\prime}$ denotes the action that is being selected based on the normalised Q-Value matrices within the global scope. The greater the Q-Value is, the greater possibility of the corresponding element inheriting an elite fragment of the current state.

*Local search*

The local search operation focuses on the adjustment of individual state. The generation of the new state is represented as follows:

$$s_{new}^j(x_1, x_2) = s_{old}^j(x_1, x_2) + \omega \times \left(s_{old}^{x_ij}(x_1, x_2) - s_{neighbor}^{x_ij}(x_1, x_2)\right) \tag{30}$$

Where $s_{new}^j(x_1, x_2)$ denotes the new state, $s_{old}^j(x_1, x_2)$ denotes the current state, and $s_{neighbor}^{x_ij}(x_1, x_2)$ denotes the neighbourhood state. $\omega$ is a random value between -1 and 1.

By implementing a combined global and local search approach, the agent and corresponding Q-Value matrices will be updated accordingly. The iteration stops once the error of the Q-value matrices generated by two adjacent actions is less than $1 \times 10^{-8}$. The state with the greatest reward among all agents is defined as the optimal solution of MPLP. The implementation procedure of HQM is shown in Code Listing 2.

---

**Algorithm 2:** Hybrid Q-Learning algorithm-based method (HQM)

---

Input: Number of lockers $M$; All task list $TK$; Number of agent $Agent$; Timestep $T$
output: The optimal state $s = (x_1, x_2)$

1 *Initialise learning rate $\alpha$, discount factor $\gamma$, random value $\sigma, \omega$, greedy factor $\epsilon$, Q-value matrices $Q_{x_1}, Q_{x_2}$*

2 **for** $i$ **in** range($Agent$) **do** // Initialise agents

3      $x_1 = $ random. randint($0, M, TK$)

4      $x_2 = $ random. permutation($TK$)

5      Combine $x_1$ and $x_2$ as a state and run ***Algorithm 1*** // Generate initial state

6 **end**

7 Rank and record states based on the reward obtained

8 **for** $t$ **in** range($T$) **do**

9      Record the best state and the corresponding reward of the agent

10      **for** $tk$ **in** range($TK$) **do**

11          Normalise Q-Value matrices $Q_{x_1}, Q_{x_2}$

12      **end**

13      **for** $i$ **in** range($Agent$) **do** // Global search

14          **for** $tk$ **in** range($TK$) **do**

15              Implement global search based on Equation (29) and obtain new variable $x_1', x_2'$

16              Combine $x_1'$ and $x_2'$ as new state $s' = (x_1', x_2')$ and run ***Algorithm 1***

17          **end**

18          Record the reward obtained by state $s' = (x_1', x_2')$ as *new_reward*

19          **if** *new_reward > current_reward* **then**

20              Update the best state and reward as $s' = (x_1', x_2')$ and *new_reward* respectively

21          **end**

22      **end**

23      **for** $i$ **in** range($Agent$) **do** // Local search

24          Implement global search based on Equation (30) and obtain new variable $x_1'', x_2''$

25          Combine $x_1''$ and $x_2''$ as new state $s'' = (x_1'', x_2'')$ and run ***Algorithm 1***

26          Record the reward obtained by state $s'' = (x_1'', x_2'')$ as *new_reward'*

27          **if** *new_reward' > new_reward* **then**

28              Update the best state and reward as $s'' = (x_1'', x_2'')$ and *new_reward'*

29          **end**

30      **end**

31      Update Q-Value matrices $Q_{x_1}, Q_{x_2}$ according to Equation (28)

32      **if** *satisfy convergence constraints* **then**

33          **break**

34      **end**

35 **end**

36 **return** *The optimal state $s = (x_1, x_2)$*

---

**CODE LISTING 2** Pseudo Code of HQM

## RESULTS AND DISCUSSION

     In this section, we discuss the application of our HQM approach and compare its effectiveness with a Genetic Algorithm (GA). We first evaluate the optimal configuration of MPLs

based on two route adjustment strategies under two algorithms with different network sizes. Additionally, we obtain and analyse the optimisation performance of HQM. We analyse the performance of the HQM and GA based on their reward acquisition capability and convergence speed. Finally, we investigate the crucial factors influencing the service delay.

## Optimal Configuration of MPLs

Testing is based on four types of whereabout settings (the total number of locations covered by each parking space equals 5, 10, 15, 20) with two route adjustment strategies applied. The parking space set for the proposed MPLP ranges from 5 to 10. The HQM and GA code are implemented in Python 3.8.5 on an Apple M1 (3.20 GHz) processor with 16GB RAM. The corresponding HQM and GA parameters are presented in Table 1. To ensure that the results are comparable, we configured the two algorithms implementation with similar parameters, to the degree possible.

We compare the performance of HQM with GA based on four aspects: i) locker deployed; ii) total travel distance; iii) average service delay; iv) reward obtained. The implemented result of the different parameter settings based on two algorithms are presented in Table 1.

In terms of the number of locker deployed, there is no significant difference between the two algorithms in achieving optimal deployment of lockers. The difference of locker deployments between the two algorithms fluctuates within the interval of [-1, 1].

The average driving distance of two algorithms based on two route policies are 308.176km (HQM-BTD), 302.748km (HQM-HCPS), 275.740km (GA-BTD), and 272.569km (GA-HCPS), respectively. Therefore, the GA achieves a better performance than HQM when it causes to minimise driving distances, due to the GA's better local search characteristic. With the increase of parking spaces deployed and customer locations, applying the HCPS strategy saves more driving distance than BTD since the round trips between parking spaces and depots are reduced.

Concerning the average service delay, the results obtained from the two algorithms based on two route policies are 10.104min (HQM-BTD), 8.907min (HQM-HCPS), 121.029min (GA-BTD), and 113.765min (GA-HCPS), respectively. Hence, the average service delay from the approach HQM is significantly lower than GA under the two route adjustment strategies, contributing to greater reward acquisition. Additionally, the HCPS strategy reduces delays by 11.847% and 6% respectively, compared to BTD implemented by HQM and GA. The average reward of two algorithms based on two route policies are $1.901 \times 10^{-3}$ (HQM-BTD), $1.965 \times 10^{-3}$ (HQM-HCPS), $0.972 \times 10^{-3}$ (GA-BTD), and $1 \times 10^{-3}$ (GA-HCPS) respectively. The result shows that the proposed HQM algorithm combined with the HCPS strategy derives better solutions than applying the BTD strategy in two algorithms.

| Number of parking spaces | Locations of each parking space | Route adjustment policy | HQM | | | | | GA | | | | |
|---|---|---|---|---|---|---|---|---|---|---|---|---|
| | | | Locker deployed (unit) | Travel distance (km) | Average service delay (min) | Reward (× 10⁻³) | Reward Gap | Locker deployed (unit) | Travel distance (km) | Average service delay (min) | Reward (× 10⁻³) | Reward Gap |
| 5 | 5 | BTD | 9 | 115.196 | 0 | 6.242 | -0.477 | 12 | 107.922 | 77.583 | 5.331 | -0.780 |
| | | HCPS | 9 | 103.842 | 0 | 6.719 | | 15 | 125.609 | 0 | 6.111 | |
| | 10 | BTD | 13 | 188.929 | 0 | 3.938 | 0.078 | 16 | 168.477 | 111.319 | 1.438 | 0.227 |
| | | HCPS | 16 | 179.039 | 0 | 3.860 | | 16 | 175.058 | 26.944 | 1.211 | |
| | 15 | BTD | 15 | 203.999 | 0 | 3.584 | -0.142 | 16 | 178.342 | 49.013 | 0.670 | -0.080 |
| | | HCPS | 16 | 184.857 | 0.044 | 3.726 | | 14 | 162.803 | 158.607 | 0.750 | |
| | 20 | BTD | 17 | 230.168 | 5.706 | 1.249 | 0.182 | 16 | 199.032 | 143.638 | 0.435 | 0.099 |
| | | HCPS | 15 | 213.614 | 8.653 | 1.067 | | 15 | 216.142 | 151.553 | 0.336 | |
| | Mean value (BTD) | | 14 | 184.573 | 1.427 | 3.753 | - | 15 | 163.443 | 95.388 | 1.969 | - |
| | Mean value (HCPS) | | 14 | 170.338 | 2.174 | 3.843 | - | 15 | 169.903 | 84.276 | 2.102 | - |
| 6 | 5 | BTD | 11 | 148.131 | 0 | 4.923 | -0.451 | 11 | 135.441 | 109.518 | 4.489 | -0.062 |
| | | HCPS | 11 | 131.085 | 0 | 5.374 | | 11 | 127.910 | 81.355 | 4.551 | |
| | 10 | BTD | 15 | 214.336 | 1.680 | 2.407 | -0.580 | 17 | 205.476 | 78.547 | 0.480 | -0.175 |
| | | HCPS | 16 | 234.621 | 0.250 | 2.987 | | 14 | 196.877 | 121.986 | 0.655 | |
| | 15 | BTD | 18 | 268.043 | 1.378 | 2.075 | 0.103 | 17 | 227.272 | 163.194 | 0.676 | -0.009 |
| | | HCPS | 17 | 260.924 | 3.788 | 1.972 | | 17 | 214.830 | 115.688 | 0.685 | |
| | 20 | BTD | 23 | 274.511 | 0.991 | 1.986 | -0.481 | 21 | 277.209 | 102.267 | 0.792 | 0.189 |
| | | HCPS | 20 | 283.338 | 0.220 | 2.467 | | 21 | 261.223 | 91.843 | 0.603 | |
| | Mean value (BTD) | | 17 | 226.255 | 1.012 | 2.848 | - | 17 | 211.350 | 113.382 | 1.609 | - |
| | Mean value (HCPS) | | 16 | 227.492 | 1.065 | 3.200 | - | 16 | 200.210 | 102.718 | 1.624 | - |
| 7 | 5 | BTD | 12 | 179.269 | 0 | 4.179 | -0.204 | 12 | 157.566 | 191.808 | 3.775 | -0.058 |
| | | HCPS | 12 | 168.145 | 0 | 4.383 | | 15 | 170.794 | 33.413 | 3.833 | |
| | 10 | BTD | 20 | 271.938 | 11.210 | 0.670 | -0.034 | 19 | 236.719 | 93.363 | 0.246 | -0.011 |
| | | HCPS | 19 | 285.177 | 10.942 | 0.704 | | 18 | 224.509 | 141.294 | 0.257 | |
| | 15 | BTD | 23 | 345.110 | 15.170 | 0.454 | 0.055 | 23 | 281.598 | 168.470 | 0.229 | 0.068 |
| | | HCPS | 25 | 337.984 | 16.348 | 0.399 | | 23 | 279.518 | 145.034 | 0.161 | |
| | 20 | BTD | 23 | 371.527 | 17.891 | 0.393 | -0.125 | 25 | 346.943 | 108.664 | 0.248 | 0.002 |
| | | HCPS | 28 | 475.139 | 10.100 | 0.518 | | 24 | 373.877 | 120.663 | 0.246 | |
| | Mean value (BTD) | | 20 | 291.961 | 11.068 | 1.424 | - | 20 | 255.707 | 140.576 | 1.125 | - |
| | Mean value (HCPS) | | 21 | 316.611 | 9.348 | 1.501 | - | 20 | 262.175 | 110.101 | 1.124 | - |
| 8 | 5 | BTD | 15 | 188.846 | 0 | 3.790 | -0.075 | 16 | 163.958 | 73.506 | 1.297 | 0.341 |
| | | HCPS | 15 | 183.766 | 0 | 3.865 | | 15 | 174.367 | 103.687 | 0.956 | |
| | 10 | BTD | 19 | 315.499 | 11.590 | 0.662 | -0.110 | 19 | 248.248 | 172.337 | 0.320 | 0.063 |
| | | HCPS | 19 | 302.573 | 9.458 | 0.772 | | 19 | 251.514 | 154.026 | 0.257 | |
| | 15 | BTD | 22 | 379.610 | 24.673 | 0.312 | -0.083 | 24 | 361.005 | 85.242 | 0.206 | -0.086 |
| | | HCPS | 23 | 364.222 | 17.830 | 0.395 | | 23 | 333.354 | 144.870 | 0.292 | |
| | 20 | BTD | 32 | 459.328 | 20.959 | 0.252 | -0.024 | 30 | 414.339 | 135.767 | 0.152 | 0.017 |
| | | HCPS | 33 | 454.509 | 18.218 | 0.276 | | 31 | 406.337 | 113.394 | 0.135 | |

| | | | | | | | | | | | |
|---|---|---|---|---|---|---|---|---|---|---|---|
| | Mean value (BTD) | | 22 | 335.821 | 14.306 | 1.254 | - | 23 | 296.888 | 116.713 | 0.494 | - |
| | Mean value (HCPS) | | 23 | 326.268 | 11.377 | 1.327 | - | 22 | 291.393 | 128.994 | 0.410 | - |
| 9 | 5 | BTD | 15 | 202.043 | 0 | 3.610 | 0.264 | 15 | 192.197 | 76.660 | 1.091 | 0.163 |
| | | HCPS | 15 | 205.974 | 0.240 | 3.346 | | 16 | 194.847 | 69.625 | 0.928 | |
| | 10 | BTD | 22 | 378.835 | 23.700 | 0.323 | -0.004 | 22 | 360.514 | 153.209 | 0.136 | -0.021 |
| | | HCPS | 22 | 388.371 | 23.241 | 0.327 | | 21 | 329.978 | 243.752 | 0.157 | |
| | 15 | BTD | 29 | 432.421 | 6.052 | 0.687 | -0.044 | 30 | 411.984 | 24.620 | 0.262 | 0.008 |
| | | HCPS | 31 | 410.161 | 5.174 | 0.731 | | 28 | 387.925 | 67.168 | 0.254 | |
| | 20 | BTD | 32 | 480.981 | 21.338 | 0.247 | -0.054 | 32 | 450.828 | 122.241 | 0.130 | -0.088 |
| | | HCPS | 30 | 460.842 | 18.043 | 0.301 | | 30 | 419.316 | 114.243 | 0.218 | |
| | Mean value (BTD) | | 25 | 373.570 | 12.773 | 1.217 | - | 25 | 353.881 | 94.183 | 0.405 | - |
| | Mean value (HCPS) | | 25 | 366.337 | 11.675 | 1.176 | - | 24 | 333.017 | 123.697 | 0.389 | - |
| 10 | 5 | BTD | 18 | 254.706 | 0 | 2.901 | 0.725 | 16 | 204.258 | 143.019 | 0.480 | -0.531 |
| | | HCPS | 19 | 249.439 | 1.211 | 2.176 | | 16 | 217.141 | 71.606 | 1.011 | |
| | 10 | BTD | 24 | 420.902 | 20.396 | 0.335 | -0.050 | 24 | 345.720 | 198.750 | 0.176 | 0.030 |
| | | HCPS | 23 | 383.965 | 18.270 | 0.385 | | 23 | 335.836 | 138.435 | 0.146 | |
| | 15 | BTD | 31 | 491.675 | 18.826 | 0.281 | 0.012 | 29 | 435.426 | 112.783 | 0.156 | -0.008 |
| | | HCPS | 31 | 461.697 | 20.048 | 0.269 | | 30 | 446.390 | 147.523 | 0.164 | |
| | 20 | BTD | 34 | 580.228 | 40.929 | 0.130 | -0.017 | 31 | 507.270 | 209.165 | 0.094 | -0.005 |
| | | HCPS | 32 | 542.657 | 31.684 | 0.147 | | 32 | 515.494 | 173.650 | 0.099 | |
| | Mean value (BTD) | | 27 | 436.878 | 20.038 | 0.912 | - | 25 | 373.169 | 165.929 | 0.227 | - |
| | Mean value (HCPS) | | 27 | 409.440 | 17.803 | 0.744 | - | 26 | 378.715 | 132.804 | 0.355 | - |

**Notes:** 1.The comparison test was implemented based on the following parameters: a) service radius = 5km; b) quantity range of locations for each customer = [1, 4]; c) working hours = 9:00-18:00; d) demand range of customer = [1,4]; e) walking speed = 0.08km/min; f) customer walking range (km) = [0.1, 1]); g) range of time span of customer (min) = [30,90]; h) range of available parking time = [30, 70]; i) locker speed = 0.7 km/hr; j) capacity of locker = 20 units.

2. HQM parameters: a) weights of the objective function $W_1 : W_2 = 5 : 1$; b) learning rate $\alpha = 0.9 \times e^{-\left(\frac{current\ iteration}{total\ iteration}\right)}$; c) discount factor $\gamma = 0.9$; d) greedy factor $\varepsilon$ follow the continuous uniform distribution $U(0,1)$; e) episode/iteration = 1000; f) population size = 100; g) random variable $\sigma$ and $\omega$ follow $U(0,1)$.

2. GA parameters: a) elite size = 5%; b) possibility of crossover = 50%; c) mutation rate = 5%.

3. Abbreviations: a) BTD = Back to Depot; b) HCPS = Holding at Current Parking Space.

4. Improvement rate = (final best reward –best initial reward) / best initial reward ×100%.

5. Reward gap = Reward (BTD) – Reward (HCPS).

6. The mean value of the number of lockers deployed is rounded upward.

**Optimisation Performance**

We further investigated concrete improvements in rewards acquired through implementing HQM and GA with two route adjustment strategies. The improvement rate is calculated based on $(Reward(final) - Reward(initial))/Reward(initial)$. The broader gap between the two surfaces in Figure 4b indicates that the HCPS obtains a more significant improvement in reward acquisition than BTD (Figure 4a) when implementing HQM. Specifically, HCPS achieves an average of 453.58% improvement in final rewards than initial rewards, while BTD achieved an average of 433.23% improvement. The average improvement of the two strategies is 443.41%.

Compared with HQM, the GA performs poorly when it comes to the route adjustment strategies since the gap between the two surfaces became significantly narrower in Figure 4c and 4d. In this case, HCPS achieves an average of 97.308% improvement in final rewards compared to initial rewards, while BTD achieved an average of 92.504% improvement. On average, the optimisation ability of HQM is 348.49% greater than that of GA, suggesting HQM enables a more efficient search for better solutions.

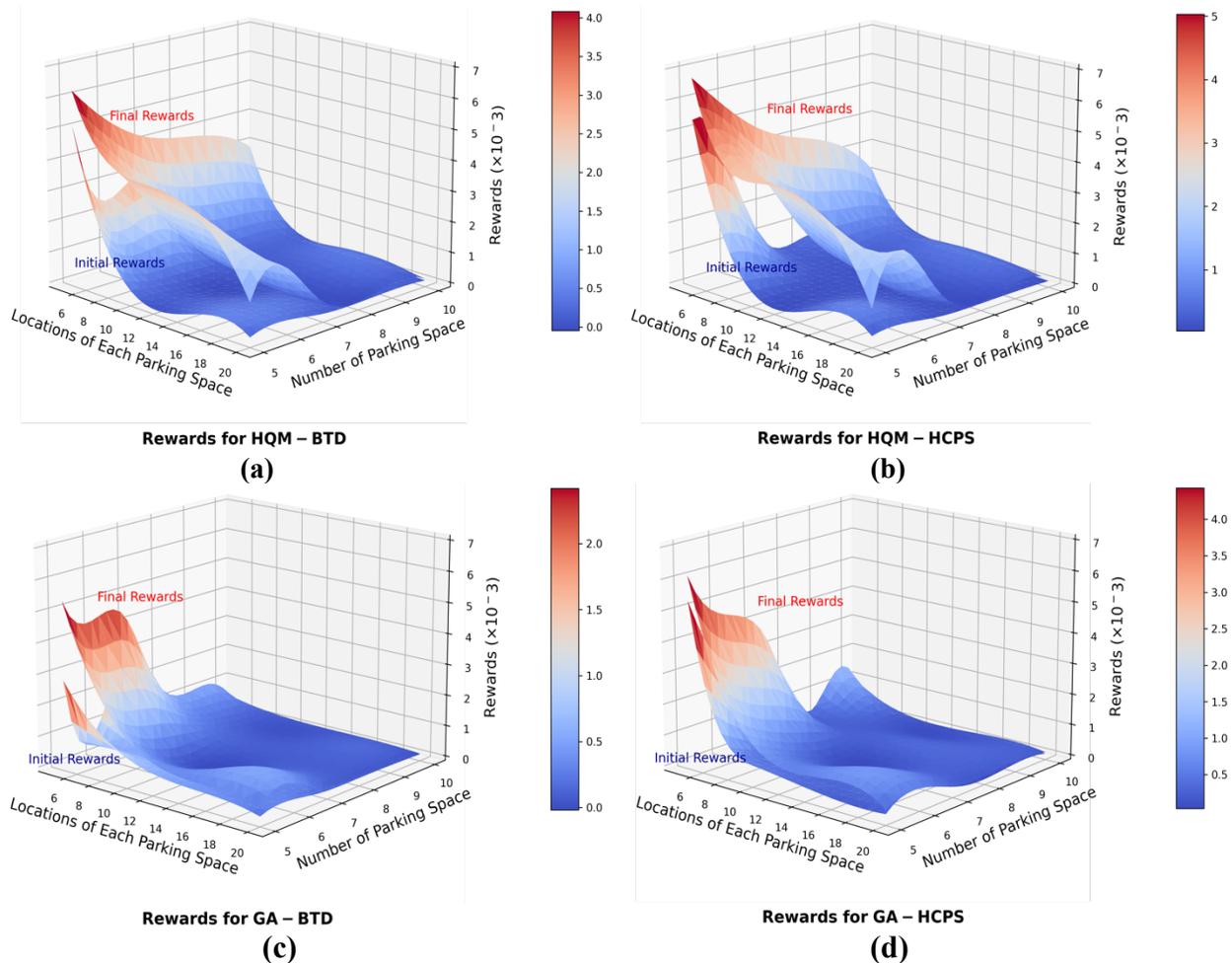

**FIGURE 4** Optimisation Ability of HQM and GA

**Reward Acquisition and Convergence Comparison**

Figure 5 shows the reward acquisition and convergence of HQM and GA under the two route adjustment policies. Algorithms are implemented under a relatively large instance setting (Number of parking spaces = 10, Number of locations of each parking space = 20). Figure 5a indicates that HQM achieves greater final rewards (BTD=$1.499 \times 10^{-3}$; HCPS=$1.373 \times 10^{-3}$) than GA (BTD=$1.188 \times 10^{-3}$; $1.076 \times 10^{-3}$) under two strategies. Furthermore, HQM steadily obtains a better reward through iteration, while GA tends to fall into local optimum during earlier iteration, resulting in a lower reward. This can be explained by Figure 5b, which capture the convergence performance of algorithms: the GA's value of objective function converged sharply at generation 200 (±20), whilst HQM iterates relatively smoothly and converged at generation 250 (±50) even though there are fluctuations in the subsequent iterations. The reason for this phenomenon is that, compared with GA, HQM makes better utilisation of the information feedback obtained by performing the previous action, to retain the elements that make a significant contribution to the optimal solution in the previous state, thus providing potential for obtaining better solutions.

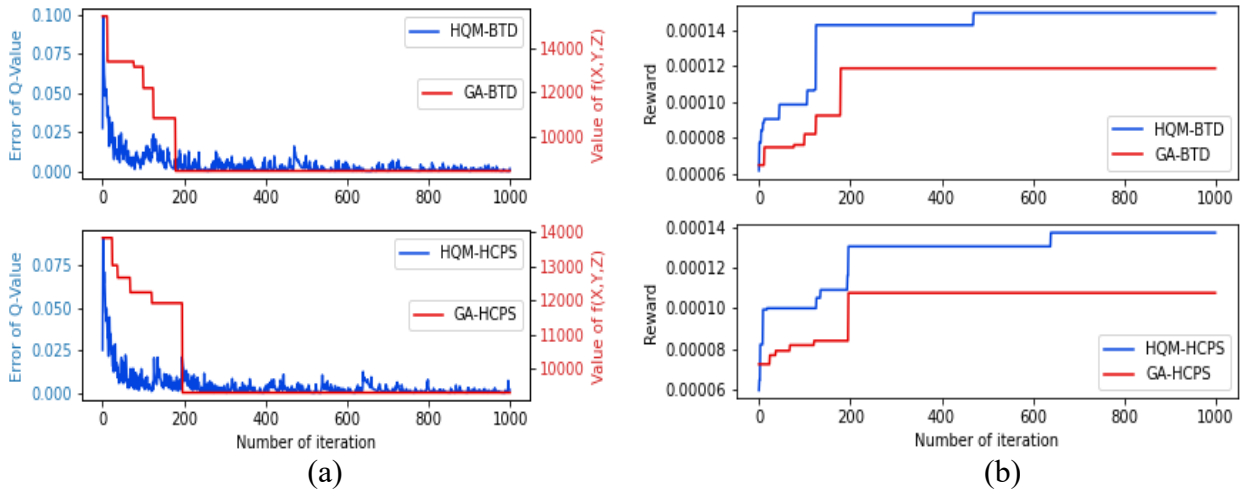

(a)                                                              (b)

**FIGURE 5** Reward Acquisition and Convergence Comparison

### Effects of Critical Factors against Service Delay

The study finally investigated how critical factors affect service delay and provide operational and managerial insight for logistics and administration professionals. Critical factors are categorised into four dimensions, depending on the attributes of factors: i) time windows from customer and parking space perspective, ii) locker properties, iii) network density, and iv) locker service radius and customer maximum walking range. The effects of different factors are summarised in Figure 6. They suggest the following findings:

Figure 6a demonstrates how customers' time windows and parking spaces' time windows affect service delays. To be more realistic, the customer residence time and available parking time are generated randomly following a normal distribution interval with a median value of $T_s$ and $T_p$, namely as $[T_s - 20, T_s + 20]$, and $[T_p - 20, T_p + 20]$ respectively. The result shows that the service delay decreased with customer residence time increased, meaning that if customers provide more generous time windows, the locker can be more flexibly scheduled, thus reducing delays. Specifically, the median threshold ($T_s$) of customer residence time is 50, corresponding to customer time windows' distribution [30,80], and the delay will be significantly reduced once $T_s \geq$ 50. Regarding the available parking time, it suffers fluctuation with $T_p = 55$. However, with the increase of $T_s$, the effect of available parking time $T_p$ on delay will gradually decrease. Based on the above analysis, customer stay time $T_s$ has more significant impacts on service delay compared to available parking time $T_p$, and the delay will decrease as $T_s$ and $T_p$ increase.

Figure 6b indicates the effect of locker attributes against delay. Generally, service delay will be improved as locker speed $V_l$ increased or locker capacity $Q$ decreased. The reduced locker capacity means fewer customers can be fulfilled in a single delivery round, which saves more travel distance, thus reducing the potential delays. However, more lockers are required to satisfy the remaining demands.

Conversely, the greater capacity may cause the algorithm to assign more customers to the same locker, reducing the number of lockers deployed and obtaining greater rewards, but may lead to higher service delays. As locker speed increases, the delay tends to be improved. However, if the moving speed is too fast, the travel time between two adjacent parking spaces may be too short for the available parking time restriction. In this case, the lockers have to adopt route adjustment strategies to resolve time windows conflicts, resulting in higher delays. Based on the above analysis, the combination of $Q = 25(\pm 5)$ unit and $V_l = 0.7 km/h$ may achieve satisfactory delay improvement.

Figure 6c shows that the parking spaces and customers' locations have conspicuous impacts on service delay, proving from the steeply curved surface. The increase of parking spaces and customer locations leads to more locker deployments and longer travel distances, resulting in higher delays. Finally, Figure 6d presents that the greater locker service radius or customer walking distance will cause higher delay since more customers should be covered within the service radius. Under this circumstance, the combination of $\rho_l = 5.5(\pm 0.5)$ $km$ and $\rho_c = 0.6(\pm 0.1)$ $km$ may obtain lower delays.

In summary, the time windows constraints and network density are still the main factors affecting service delay. This requires coordination between logistics professionals, customers, and urban planners to determine a better combination of critical factors to improve timeliness more economically. Additionally, for carriers, defining a reasonable service radius and effective scheduling policies based on customer preferences will increase customer satisfaction.

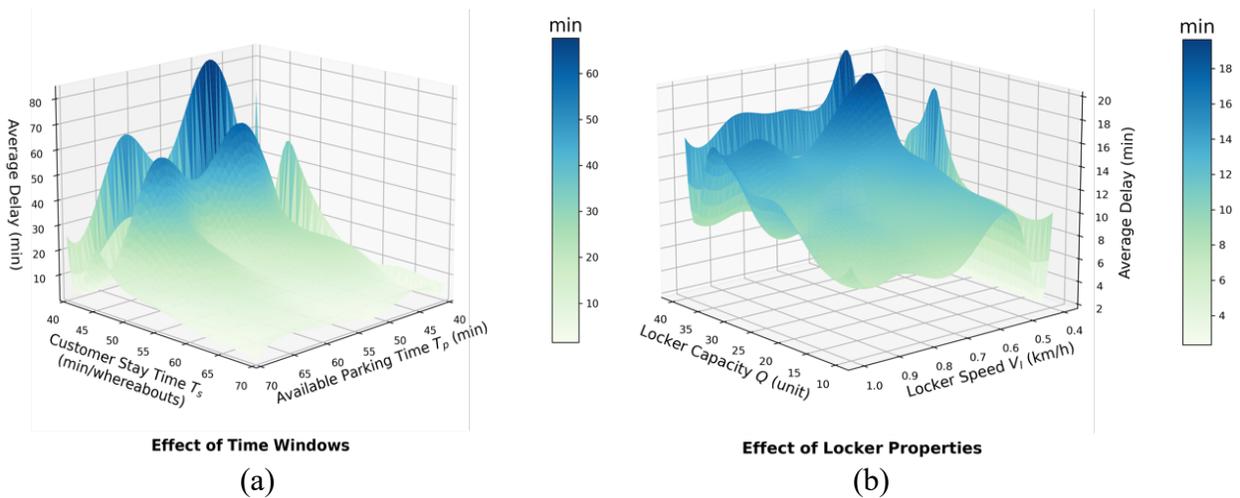

**Effect of Time Windows**

(a)

**Effect of Locker Properties**

(b)

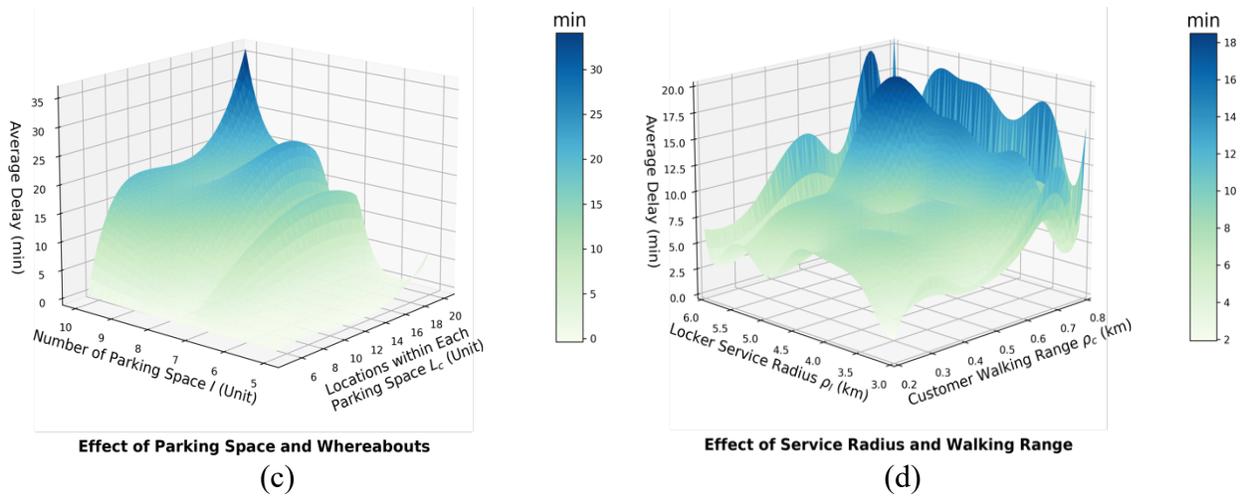

(c)                 (d)

**FIGURE 6** Effects of Critical Factors against Service Delay

## CONCLUSIONS

This study provides a novel approach for MPLP, considering customer locations and stochastic events. A hybrid Q-Learning algorithm that integrates both global and local search mechanisms has been developed to optimise locker deployed and service delays. Two route adjustment strategies are designed to resolve stochastic events. To assess the performance of the HQM, the study proposes a GA as a baseline case and tests the approach under different network density and route adjustment strategies.

The results demonstrate that our proposed HQM combined with HCPS strategy achieves better locker configuration and service delay reduction than BTD counterparts. The larger gap region of the optimisation performance surface plot shows that HQM has better optimisation ability than GA, especially in the service delay improvement aspect. It demonstrates that the proposed HQM can improve the local optimum issues faced by most evolutionary algorithms.

This study provides a new scope to solve the MPLP, which can be applied to urban deliveries with varied scales. Noticeably, it provides a novel and effective approach to flexibly schedule the autonomous delivered vehicles to urban logistics in the future.

Further work will focus on optimising more complex situations by introducing dynamic cases, where more flexible route adjustment approaches are developed in response to demand changes and delay issues. It would also be beneficial to investigate the multi-echelon MPLP, in which transhipment centres or feeder vehicles are introduced to shorten the travel distance between lockers and the central depot.

## ACKNOWLEDGMENTS

I would like to express my sincere gratitude to my supervisor and all co-authors, for their pertinent suggestions and encouragement during my study and paper creation. Specially, thanks to Miss. Xu Fang for her encouragement and support.

## AUTHOR CONTRIBUTIONS

The authors confirm contribution to the paper as follows: Conceptualisation, algorithm design, writing: Yubin Liu; Data visualisation, data curation: Qiming Ye; Methodology review, code review: Yuxiang Feng; Editing, manuscript review: Jose Escribano-Macias; Supervision, review